\documentclass[letterpaper]{article} 
\usepackage[preprint]{aaai2027}  
\usepackage[hyphens]{url}  
\usepackage{graphicx} 
\usepackage{natbib}  
\usepackage{caption} 
\usepackage{subcaption}
\usepackage{amsmath}
\usepackage{amssymb}
\usepackage{booktabs}
\usepackage{multirow}
\usepackage{xcolor}
\usepackage{xspace}
\usepackage[table]{xcolor}
\usepackage{microtype}

\newcommand{\approach}{ParaTempo\xspace}

\title{\approach: Efficient Parallel Reasoning via Temporal Confidence}
\author{
  Xuteng Zhang\textsuperscript{\rm 1}\equalcontrib, \quad
  Wenhao Zeng\textsuperscript{\rm 1}\equalcontrib, \quad
  Xiaodong Gu\textsuperscript{\rm 1}, \quad
  Chao Hu\textsuperscript{\rm 1}, \quad
  Haotian Lin\textsuperscript{\rm 1}, \quad
  Yuling Shi\textsuperscript{\rm 1}, \quad
  Min Wang\textsuperscript{\rm 2}, \quad
  Beijun Shen\textsuperscript{\rm 1}\corresponding
}
\affiliations{
  \textsuperscript{\rm 1}Shanghai Jiao Tong University \quad
  \textsuperscript{\rm 2}University of Pennsylvania \\
  \{scottzhang, zengwh\_cs, xiaodong.gu\}@sjtu.edu.cn
}

\begin{document}

\maketitle

\begin{abstract}
Parallel reasoning improves the accuracy and robustness of large reasoning models by exploring multiple solution paths, but its computational cost grows with reasoning depth and branch count. Existing methods for managing these parallel paths typically rely on final-answer consensus, local token confidence, or isolated intermediate probes. However, these signals are often delayed, weakly tied to actual reasoning progress, or too noisy for dynamic, branch-level control. To address these limitations, we introduce \approach, a training-free asynchronous parallel reasoning framework. \approach is driven by \textit{temporal confidence}, a branch-local measure of answer-space convergence. Each branch is periodically probed for a tentative answer probability distribution, and temporal confidence quantifies how sharply the recent intermediate probes concentrate on a dominant answer. Once sufficient evidence has accumulated, \approach drives its entire control process from this single signal: low-confidence branches are pruned, branches that persistently commit to their dominant answer are retired early, freed computation is reallocated by forking new branches, and generation stops globally once the confidence-weighted vote concentrates. Without requiring synchronization among reasoning trajectories, \approach adaptively allocates computation based on branch-level convergence.
Experiments on challenging mathematical and scientific reasoning benchmarks show that \approach reduces average latency by 21.8--32.2\% and total token usage by 18.1--30.3\% while maintaining competitive accuracy. Moreover, temporal confidence exhibits stronger temporal stability and predictive power for future branch convergence than token-level and instantaneous signals
\footnote{Code and dataset are available at \url{https://github.com/ScottZhang812/ParaTempo}.}.
\end{abstract}

\section{Introduction}

Parallel reasoning has emerged as an effective strategy for improving the reliability of large reasoning models by concurrently exploring multiple solution trajectories~\citep{wei2022chainofthought,wang2022selfconsistency}. 
However, its inference cost scales with both the number of branches and reasoning depth. 
Most traditional approaches \cite{wang2022selfconsistency} follow a fixed-budget paradigm, allocating identical computation to all branches and aggregating answers only after termination, thereby overlooking heterogeneous reasoning progress across trajectories. 
As a result, converged branches continue generating redundant tokens, while less promising branches consume computation without clear evidence of future utility.

\begin{figure}[t]
    \centering
    \includegraphics[width=\columnwidth]{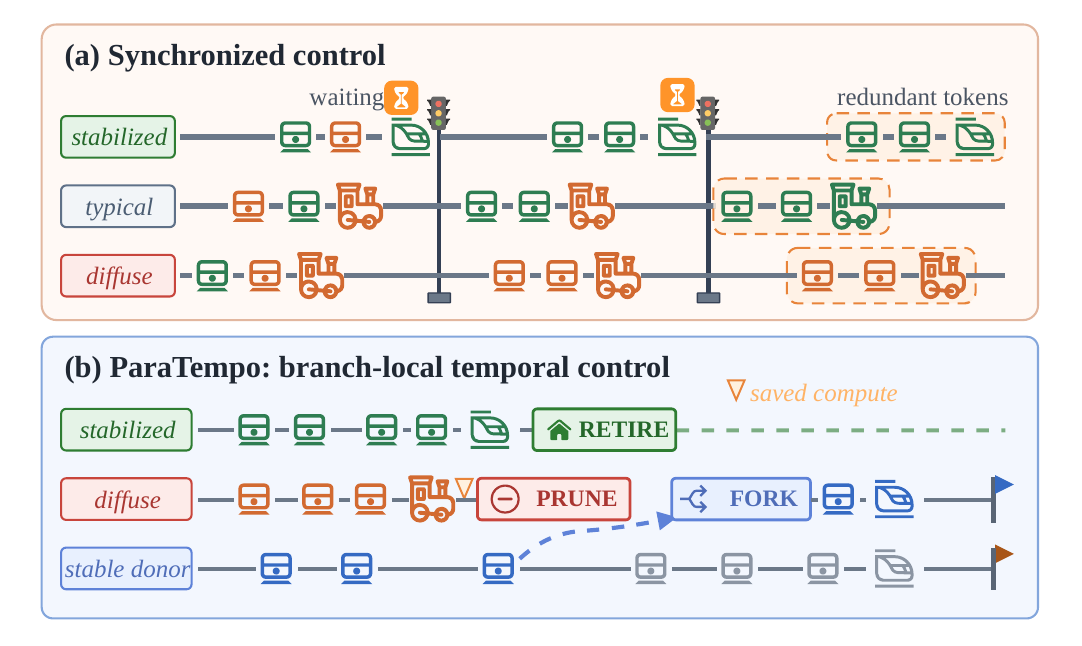}
    \caption{Motivating branch-local temporal control. (a) Synchronized control advances all branches through global barriers. (b) \approach asynchronously updates branch-local evidence and triggers local actions: stabilized branches are retired, diffuse branches are pruned, and reclaimed budget is reallocated via forking. }
    \label{fig:intro-motivation}
\end{figure}

To improve computation efficiency, recent studies have explored online control mechanisms that intervene reasoning branchs during generation~\cite{li2024esc,liu2025answerconvergence,fu2025deepconf,zheng2026parallelprobe}. For instance, early-stopping self-consistency methods reduce sampling once final-answer consensus becomes sufficiently reliable \citep{li2024esc}. Answer-convergence methods leverage the observation that many reasoning traces stabilize before generation terminates \citep{liu2025answerconvergence}. DeepConf uses model confidence to filter low-quality traces during or after generation \citep{fu2025deepconf}. 
However, these approaches rely on signals that only partially capture branch-level reasoning progress. 
Final-answer consensus offers reliable evidence but emerges only after substantial computation~\cite{li2024esc,wan2025rasc,zhou2025rpc}; token-level confidence provides fine-grained feedback but poorly reflects answer evolution~\cite{fu2025deepconf,zeng2025pruning,zeng2026glimprouter}; and isolated intermediate probes enable early intervention but are sensitive to transient fluctuations~\cite{liu2025answerconvergence,zheng2026parallelprobe}.

To overcome these limitations, we propose \approach, an asynchronous framework for efficient parallel reasoning. At its core, \approach relies on \textit{temporal confidence}, a branch-local signal for selecting and pruning branches. Temporal confidence is designed to answer a fundamental question: does a branch consistently place probability mass on a concentrated answer region, or scatter mass across multiple alternatives? To estimate this signal, \approach periodically probes each active branch to obtain tentative answer distributions and aggregates recent predictions within a sliding window. By continuously tracking the temporal consistency of intermediate predictions, temporal confidence determines whether a trajectory is stabilizing toward a dominant answer or remains uncertain among competing candidates. Guided by this signal, \approach dynamically allocates computation: low-confidence branches are pruned, converged branches are retired, freed computation is reallocated through forking from promising branches, and inference terminates when the confidence-weighted ensemble vote becomes sufficiently concentrated. 
As illustrated in Figure~\ref{fig:intro-motivation}, \approach enables asynchronous branch-level control without requiring trajectory synchronization, reducing both total generation cost and critical-path latency.

We evaluate \approach on four mathematical and scientific reasoning benchmarks using Qwen3.5-35B-A3B~\citep{qwen3.5} and GPT-OSS-20B~\citep{openai2025gptoss120bgptoss20bmodel}. 
Across these benchmarks, \approach consistently improves inference efficiency while preserving the accuracy gains of parallel reasoning. 
Compared with standard self-consistency, \approach reduces average latency by 21.8--32.2\% and total generated tokens by 18.1--30.3\% while maintaining competitive accuracy.
Compared with adaptive parallel-reasoning baselines, \approach achieves comparable or higher accuracy with substantially lower latency, and outperforms the strongest parallel controller by 3.8--3.9 accuracy points.
Further analysis confirms that temporal confidence provides a more stable and predictive indicator of future answer stability than token-level and instantaneous signals.

Our contributions are summarized as follows:
\begin{itemize}
    \item We introduce \textit{temporal confidence}, a branch-local convergence signal derived from temporally aggregated intermediate answer distributions for tracking answer stability during reasoning.

    \item We propose \approach, an asynchronous parallel reasoning framework that leverages temporal confidence for adaptive computation allocation, including branch pruning, retirement, forking, and global early termination.

    \item We conduct extensive experiments on challenging reasoning benchmarks, demonstrating that \approach substantially reduces inference cost while maintaining competitive reasoning accuracy.
\end{itemize}

\section{Problem Formulation}
Given an input problem $x$ and a model $M$, parallel reasoning generates $K$ independent reasoning branches \(r_1,...,r_K\), where each branch \(r_i\) maintains a reasoning prefix $r_{i,t}$ at time step \(t\) and eventually produces an answer $y_i$. 
Traditional methods allocate the same generation budget to all branches and aggregates final answers after completion. 
Although effective, this strategy ignores heterogeneous reasoning progress: some branches converge early and continue generating redundant tokens, while others remain uncertain and may benefit from additional exploration.

We formulate parallel reasoning as an online resource allocation problem. 
A controller should dynamically decide whether each branch continues decoding, terminates early, or receives additional computation from released resources, while preserving the reliability of final answer aggregation.

Formally, let $\pi$ denote a parallel reasoning controller and $\pi_{\mathrm{fixed}}$ the standard fixed-budget execution. 
Let $\mathcal{A}(\cdot)$ and $\mathcal{C}(\cdot)$ denote the expected accuracy and inference cost, respectively. 
The objective is to minimize inference cost while maintaining the accuracy achieved by fixed-budget execution:
\begin{equation}
    \label{eq:objective}
    \min_{\pi}\;\mathcal{C}(\pi)
    \quad
    \text{s.t.}
    \quad
    \mathcal{A}(\pi)\geq\mathcal{A}(\pi_{\mathrm{fixed}}),
\end{equation}

\section{A Preliminary Study}
\label{sec:prelim}
The formulated objective requires a controller to make reliable branch-level decisions during generation. 
We therefore investigate whether existing control signals provide two properties essential for such decisions: temporal stability and predictive ability for future branch convergence. 
Specifically, we evaluate representative token-level uncertainty and answer-level confidence signals under an uncontrolled parallel reasoning setting.

\subsection{Study Design}

For each problem, we generate multiple independent reasoning branches until their default termination. 
During decoding, each active branch is periodically probed every $\tau$ generated tokens to obtain a tentative answer distribution from its current reasoning prefix. 
We conduct this study on Qwen3.5-35B-A3B~\citep{qwen3.5} and GPT-OSS-20B~\citep{openai2025gptoss120bgptoss20bmodel} using AIME 2026~\citep{aime26}, HMMT November 2025, and HMMT February 2026~\citep{dekoninck2026matharena}, with $K=16$ branches per problem and $\tau=500$. 
This process collects over 90 million reasoning tokens. 
Unless otherwise specified, all branches use the same sampling configuration and generation budget as the main experiments.

\paragraph{Intermediate Answer Probing.}
At the $t$-th probe of branch $i$, we append an answer-forcing suffix (e.g., \texttt{</think> Final answer:}) to the current reasoning prefix and obtain the top-$L$ candidate answer tokens $V_{i,t}$ with their log probabilities $\ell_{i,t}(v)$. 
Candidates are mapped into normalized answer buckets according to the task format, and the resulting answer distribution is computed as:

\begin{equation}
    \label{eq:probe-dist}
    p_{i,t}(v)=
    \frac{\exp(\ell_{i,t}(v))}
    {\sum_{u\in V_{i,t}}\exp(\ell_{i,t}(u))},
    \qquad v\in V_{i,t}.
\end{equation}

\paragraph{Studied Signals.}
We examine three representative signals covering two common control paradigms: token-level uncertainty and answer-level confidence. 
For a distribution $q$, its entropy is defined as $H(q)=-\sum_v q(v)\log q(v)$.

Token-level signals characterize local generation uncertainty. 
Let $q_{i,j}$ denote the next-token distribution at the $j$-th decoding position of branch $i$, and $w_{i,j}$ the sampled token. 
For the tokens generated since the previous probe, denoted as $\mathcal{T}_{i,t}$, we consider:

Mean token entropy:
\begin{equation}
    \label{eq:token-entropy}
    \bar H^{\mathrm{tok}}_{i,t}
    =
    \frac{1}{|\mathcal{T}_{i,t}|}
    \sum_{j\in\mathcal{T}_{i,t}}H(q_{i,j}),
\end{equation}
and token perplexity:
\begin{equation}
    \label{eq:token-ppl}
    \mathrm{PPL}_{i,t}
    =
    \exp
    \Big(
    -\frac{1}{|\mathcal{T}_{i,t}|}
    \sum_{j\in\mathcal{T}_{i,t}}
    \log q_{i,j}(w_{i,j})
    \Big).
\end{equation}
The answer-level confidence is derived directly from the probe distribution:
\begin{equation}
    \label{eq:inst-signal}
    C^{\mathrm{inst}}_{i,t}
    =
    \exp(-H(p_{i,t})).
\end{equation}
Although this signal is aligned with the answer space, it only reflects a single observation of an evolving reasoning trajectory.

\paragraph{Evaluation Protocol.}
We evaluate each signal from two perspectives: temporal stability and future convergence prediction. 
Given a signal sequence $s_{i,1},\ldots,s_{i,T_i}$, its temporal volatility is measured as:

\begin{equation}
    \operatorname{Vol}(s_i)=
    \frac{1}{T_i-1}
    \sum_{t=2}^{T_i}|s_{i,t}-s_{i,t-1}|.
\end{equation}

To evaluate predictive ability, we measure whether a signal value indicates future answer consistency. 
Let $\tilde y_{i,t}=\arg\max_v p_{i,t}(v)$ denote the dominant answer at probe $t$. 
A probe state is considered stable if the dominant answer remains unchanged over the next $h$ probes:

\begin{equation}
    \operatorname{Stable}_{i,t}^{(h)}
    =
    \mathbf{1}
    \{\tilde y_{i,t+1}=\cdots=\tilde y_{i,t+h}=\tilde y_{i,t}\}.
\end{equation}

We group probe states by their signal values and report the empirical future-stability rate for each group, together with the Spearman correlation between each signal and $\operatorname{Stable}_{i,t}^{(h)}$ and a directional AUC that measures how well the signal separates stable probe states from unstable ones.

\subsection{Results and Analysis}


\begin{table}[t]
    \centering
    \small
    \setlength{\tabcolsep}{3pt}
    \begin{tabular}{lccc}
        \toprule
        \textbf{Signal}                & \textbf{Volatility} $\downarrow$ & \textbf{Spearman} $|\rho|$ $\uparrow$ & \textbf{AUC} $\uparrow$ \\
        \midrule
        Mean token entropy             & 0.54                             & 0.13                                  & 0.58                    \\
        Token perplexity               & 0.56                             & 0.12                                  & 0.57                    \\
        Inst.\ answer confidence       & 0.26                             & 0.41                                  & 0.71                    \\
        \bottomrule
    \end{tabular}
    \caption{Evaluation of existing signals by temporal volatility and future answer stability prediction ($h=5$). Spearman correlation and AUC measure signal stability and predictive performance, respectively.}
    \label{tab:prelim-signals}
\end{table}

\paragraph{Token-Level Signals Lack Answer-Space Alignment.}
Table~\ref{tab:prelim-signals} summarizes the diagnosis.
Token-level uncertainty measures, including token entropy and perplexity, exhibit the largest fluctuations across adjacent probes. 
These variations are largely driven by local linguistic factors rather than changes in the underlying answer state. 
A branch may therefore maintain low token-level uncertainty while exploring different answer hypotheses, or exhibit high lexical uncertainty despite approaching a stable solution. 
Besides, their association with future answer stability is weak ($|\rho|\leq 0.13$, AUC $\leq 0.58$), indicating that token-level signals provide limited evidence about whether further decoding is likely to improve the final answer.

\paragraph{Instantaneous Answer Confidence Is Sensitive to Transient Changes.}
Intermediate reasoning steps may temporarily favor incorrect or unstable hypotheses, causing substantial fluctuations in the probe distribution. Its standardized volatility (0.26) is not negligible. Thus, decisions based on instantaneous answer confidence can be sensitive to short-term variations and may not accurately reflect the long-term convergence behavior of a branch.

These observations reveal the limitations of existing signals for fine-grained online reasoning control. 
An effective control signal should satisfy three properties: 
(1)~\textbf{answer-space alignment}, by measuring the distribution over candidate answers rather than surface-level generation statistics; 
(2)~\textbf{temporal consistency}, by aggregating evidence across multiple observations instead of relying on a single probe; and 
(3)~\textbf{branch locality}, by enabling independent updates without requiring cross-branch synchronization.

\section{Methodology}

\begin{figure*}[t]
    \centering
    \includegraphics[width=\textwidth]{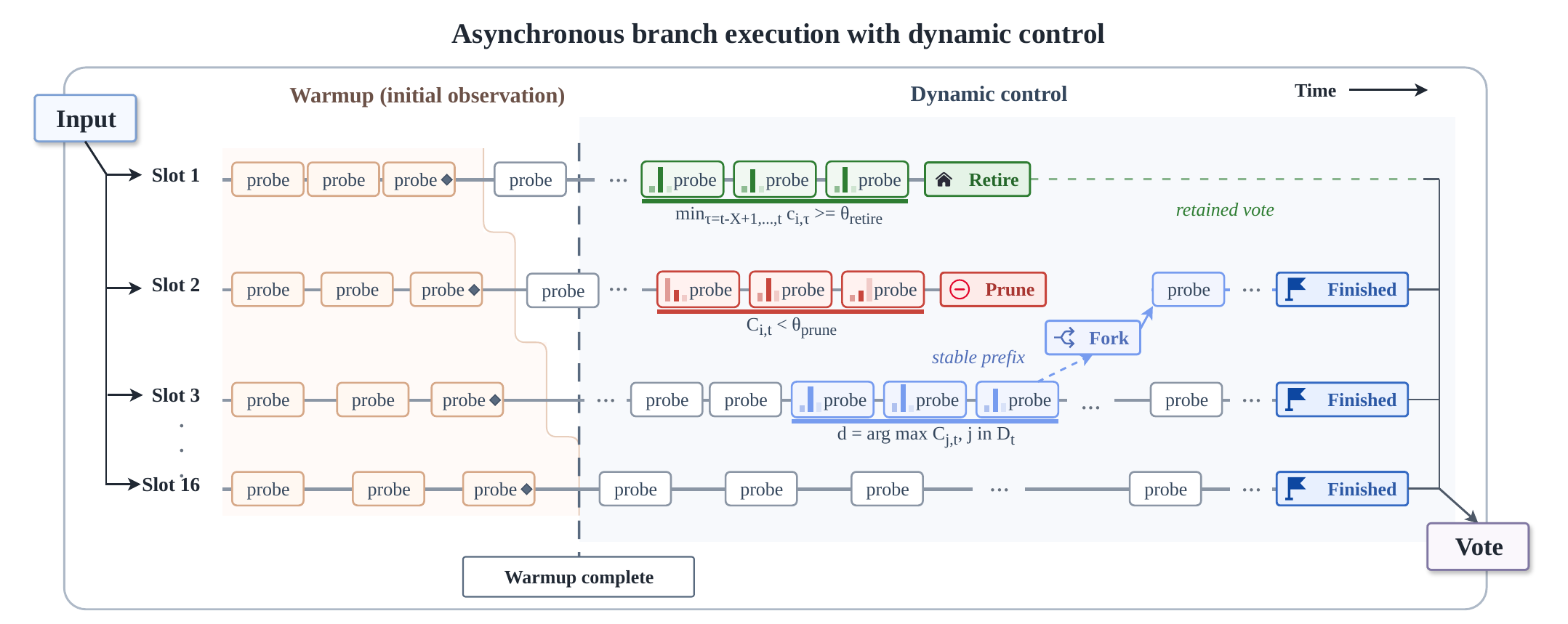}
    \caption{Overall framework of \approach. The framework periodically probes reasoning branches, estimates temporal confidence, and asynchronously allocates computation through branch control and confidence-weighted voting.}
    \label{fig:framework}
\end{figure*}

To improve the efficiency of parallel test-time reasoning, we propose \approach, a training-free asynchronous framework for branch-level computation allocation. \approach is driven by a new control signal called \textit{temporal confidence} (Section \ref{sec:tempconf}). The key idea is to convert intermediate answer evolution into an online control signal: branches with stable answer distributions are terminated early, while branches with uncertain trajectories receive additional exploration.

\subsection{Overview}

Figure~\ref{fig:framework} illustrates the overall workflow.
Given a set of sampled reasoning branches, \approach periodically probes active branches to obtain intermediate answer distributions from their current reasoning prefixes.
Recent probe distributions are aggregated to update each branch's temporal confidence, which guides asynchronous branch control through continued decoding, retirement, pruning, and forking.
The controller first performs a warmup stage to calibrate a problem-specific decision threshold and subsequently manages branch states without requiring synchronization across branches.
The final prediction is obtained by aggregating the dominant answers of branches, each weighted by its top-1 probability.

At any time, each branch is assigned one of four states: \textsc{Active}, \textsc{Retired}, \textsc{Pruned}, or \textsc{Forked}. 
Active and forked branches consume additional tokens, retired branches preserve voting evidence, and pruned branches release computation for exploration. 
This state-based design allows \approach to dynamically allocate computation while preserving sufficient reasoning diversity.

\subsection{Temporal Confidence}
\label{sec:tempconf}
Motivated by the preliminary study, we introduce \textit{temporal confidence}, an online branch-level signal that measures answer-space convergence over time. 
Given a reasoning prefix, \approach probes the model's intermediate answer distribution, aggregates recent observations, and measures the resulting answer-space concentration for adaptive computation control.

\paragraph{Temporal Aggregation.}
A single probe distribution may be noisy due to transient intermediate reasoning states. 
To reduce such fluctuations, \approach aggregates recent probe distributions within a sliding window of size $W$:
\begin{equation}
    \label{eq:agg-dist}
    g_{i,t}(v)=
    \frac{1}{|\mathcal{J}_{i,t}|}
    \sum_{\tau\in\mathcal{J}_{i,t}}p_{i,\tau}(v),
\end{equation}
where $\mathcal{J}_{i,t}=\{\max(1,t-W+1),\ldots,t\}$ denotes the probe steps included in the window. 
The support of $g_{i,t}$ is defined as the union of answer buckets observed in the window,
$\mathcal{V}_{i,t}=\bigcup_{\tau\in\mathcal{J}_{i,t}}V_{i,\tau}$, with candidates absent from a particular probe assigned zero probability. 
This temporal aggregation emphasizes persistent answer preferences while suppressing isolated probe fluctuations.

\paragraph{Confidence Estimation.}
We define \textit{temporal confidence} as the exponentiated negative entropy of the aggregated distribution:
\begin{equation}
    \label{eq:temporal-confidence}
    C_{i,t}=\exp\big(-H(g_{i,t})\big)\in(0,1].
\end{equation}
Since $\exp(H(g_{i,t}))$ corresponds to the perplexity of the aggregated answer distribution, $C_{i,t}$ can be interpreted as the inverse effective number of competing answer candidates. 
It approaches $1$ when recent probes consistently concentrate on a single answer and decreases as probability mass spreads across multiple alternatives.

In addition to temporal confidence, we track the dominant answer and its probability mass:
\begin{equation}
    \label{eq:top1}
    \hat y_{i,t}=\arg\max_{v}g_{i,t}(v),
    \qquad
    c_{i,t}=\max_{v}g_{i,t}(v).
\end{equation}

Temporal confidence captures the overall convergence state of a branch and serves as the primary signal for adaptive computation control. 
The dominant answer $\hat y_{i,t}$ provides the current branch prediction, while $c_{i,t}$ measures its support and is used for branch retirement and confidence-weighted answer aggregation.

\subsection{Asynchronous Branch Control}

Building on temporal confidence, \approach performs branch-level asynchronous computation allocation during generation. 
At each probe point, a branch independently determines whether to continue decoding, retire after convergence, or release computation for further exploration. 
Unlike synchronous parallel reasoning, these decisions do not require branches to align at the same reasoning depth, avoiding unnecessary waiting caused by heterogeneous trajectory lengths. 
By eliminating width-wise synchronization barriers, \approach reduces both total computation and critical-path latency.
The full pseudocode of the overall control procedure is provided in the Technical Supplement.

\paragraph{Warmup Calibration.}
The pruning threshold should adapt to different problems and models. \approach therefore begins with a warmup phase of $N_{\mathrm{warm}}$ probes, during which all branches generate normally and no branch-level intervention is performed.
The collected temporal-confidence values are used to calibrate an instance-specific pruning threshold:
\begin{equation}
    \mathcal{S}_{\mathrm{warm}} =
    \{C_{i,t}: i\in\{1,\ldots,K\},\; W\leq t\leq N_{\mathrm{warm}}\}.
\end{equation}
The pruning threshold is set by a quantile:
\begin{equation}
    \theta_{\mathrm{prune}} =
    \operatorname{Quantile}_{1-q_{\mathrm{prune}}}
    (\mathcal{S}_{\mathrm{warm}}).
\end{equation}
The quantile-based calibration adapts the pruning criterion to the confidence distribution of the current problem, avoiding a manually specified global threshold.

\paragraph{Branch Pruning.}
After warmup, branches with insufficient temporal confidence, i.e., $C_{i,t}<\theta_{\mathrm{prune}}$, are removed.
Such branches exhibit diffuse answer distributions, indicating that additional decoding is unlikely to provide reliable convergence.
Pruning terminates these low-value trajectories and releases computation for new exploration.
To avoid premature pruning, we require each branch to accumulate sufficient post-fork probe history before applying this criterion.

\paragraph{Early Retirement.}
A branch may achieve local convergence before the entire ensemble reaches consensus.
\approach therefore retires a branch when its dominant answer remains sufficiently concentrated over $X$ consecutive probes:
\begin{equation}
    \min_{\tau\in\{t-X+1,\ldots,t\}} c_{i,\tau}
    \geq \theta_{\mathrm{retire}}.
\end{equation}
A retired branch stops consuming generation tokens while preserving its answer $\hat y_{i,t}$ and confidence weight $c_{i,t}$ for final aggregation.
This mechanism converts early convergence into computation savings without discarding useful evidence.

\paragraph{Adaptive Forking.}
When pruning releases a computation slot, \approach reallocates the budget by forking from a promising branch to maintain exploration capacity.
Let $\mathcal{D}_t$ denote eligible donor branches with sufficient temporal confidence and probe history.
The donor is selected as:
\begin{equation}
    d = \arg\max_{j\in\mathcal{D}_t} C_{j,t}.
\end{equation}
The forked branch inherits the donor's reasoning prefix and continues with an independent sampling seed.
This strategy exploits promising partial solutions while preserving trajectory diversity.
If no eligible donor exists, the released slot remains inactive.


\subsection{Global Consensus and Answer Aggregation}
While branch-level control determines how computation is allocated, \approach uses a global consensus mechanism to decide when sufficient evidence has been accumulated and to produce the final prediction.
Let $\mathcal{B}_t$ denote the set of eligible voting branches at time $t$, including active and retired branches with valid temporal-confidence estimates.
For an answer bucket $a$, we define the confidence-weighted vote mass as:
\begin{equation}
    V_t(a)=
    \sum_{i\in\mathcal{B}_t}
    c_{i,t}\mathbf{1}\{\hat y_{i,t}=a\}.
\end{equation}
The generation process terminates early when the dominant answer receives sufficient aggregate confidence:
\begin{equation}
    \max_a V_t(a) \geq \gamma_{\mathrm{ES}}|\mathcal{B}_t|.
\end{equation}
Otherwise, \approach continues until no active branch remains or the computation budget is exhausted, and returns the final prediction:
\begin{equation}
    \hat y =
    \arg\max_a V_t(a).
\end{equation}

\section{Experiments}

\subsection{Experimental Setup}
\begin{table*}[t]
    \centering
    \small
    \setlength{\tabcolsep}{1.5pt}
        \begin{tabular}{lrrrrrrrrrrrrrrrr}
            \toprule
            \textbf{Method}   & \multicolumn{4}{c}{\textbf{AIME26}} & \multicolumn{4}{c}{\textbf{HMMT25}} & \multicolumn{4}{c}{\textbf{HMMT26}} & \multicolumn{4}{c}{\textbf{GPQA}}                                                                                                                                                                                                                                              \\
            \cmidrule(lr){2-5} \cmidrule(lr){6-9} \cmidrule(lr){10-13} \cmidrule(lr){14-17}
                              & Acc. $\uparrow$                     & Lat. $\downarrow$                   & Tok. $\downarrow$                   & Seq. $\downarrow$                 & Acc. $\uparrow$  & Lat. $\downarrow$ & Tok. $\downarrow$ & Seq. $\downarrow$ & Acc. $\uparrow$  & Lat. $\downarrow$ & Tok. $\downarrow$ & Seq. $\downarrow$ & Acc. $\uparrow$  & Lat. $\downarrow$ & Tok. $\downarrow$ & Seq. $\downarrow$ \\
            \midrule
            \multicolumn{17}{c}{\cellcolor{gray!15}\textit{Base model: Qwen3.5-35B-A3B}}                                                                                                                                                                                                                                                                                                                                         \\
            \midrule
            Zero-shot         & 72.3                                & 92.2                                & 12.6k                               & 12.6k                             & 64.2             & 90.6              & 12.4k             & 12.4k             & 42.4             & 94.7              & 13.0k             & 13.0k             & 82.2             & 69.0              & 9.4k              & 9.4k              \\
            SC@16             & 87.5                                & 250.6                               & 229.7k                              & 15.6k                             & 69.2             & 257.8             & 236.8k            & 16.0k             & 45.5             & 254.8             & 237.4k            & 15.8k             & 86.4             & 225.0             & 196.7k            & 14.7k             \\
            ESC@16            & 83.3                       & 279.7                               & 105.4k                              & 27.2k                             & 70.0 & 333.1             & 125.9k            & 32.3k             & 42.4    & 369.8             & 140.0k            & 35.9k             & 87.0    & 218.6             & 79.6k             & 21.6k             \\
            SAC@16            & 73.3                                & 216.9                               & 144.8k                              & 12.8k                             & 68.3             & 217.2             & 141.9k            & 13.4k             & 34.8 & 216.9             & 141.6k            & 13.4k             & 81.8             & 144.8    & 102.8k            & 9.1k              \\
            DeepConf-high@16  & 68.3                                & 451.4                               & 101.3k                              & 101.3k                            & 60.0             & 433.1             & 99.6k             & 99.6k             & 34.8 & 460.6             & 102.5k            & 102.5k            & 82.3             & 271.8             & 79.2k             & 79.2k             \\
            DeepConf-low@16   & 65.0                                & 190.0                      & 102.9k                              & 102.9k                            & 53.3             & 190.9    & 103.7k            & 103.7k            & 31.8             & 200.1    & 104.2k            & 104.2k            & 82.8             & 154.2 & 94.1k             & 94.1k             \\
            Parallel-Probe@16 & 76.7                    & 223.1                               & 164.0k                              & 12.5k                             & 65.0             & 218.1             & 161.2k            & 12.3k             & 42.4    & 220.3             & 161.9k            & 12.3k             & 84.6             & 203.5             & 153.7k            & 11.4k             \\
            \approach{}@16    & 83.3                       & 198.4                   & 161.9k                              & 10.2k                             & 73.3    & 205.7 & 166.6k            & 10.6k             & 42.4    & 208.0 & 168.9k            & 10.8k             & 85.4 & 161.1             & 130.4k            & 8.7k              \\
            \midrule
            \multicolumn{17}{c}{\cellcolor{gray!15}\textit{Base model: GPT-OSS-20B}}                                                                                                                                                                                                                                                                                                                                             \\
            \midrule
            Zero-shot         & 70.0                                & 40.0                                & 6.8k                                & 6.8k                              & 56.7             & 51.5              & 8.7k              & 8.7k              & 45.5             & 56.9              & 9.7k              & 9.7k              & 67.2             & 24.9              & 4.5k              & 4.5k              \\
            SC@16             & 90.0                                & 110.6                               & 119.3k                              & 11.2k                             & 68.3             & 136.8             & 148.1k            & 13.3k             & 56.8             & 144.7             & 158.0k            & 13.6k             & 72.2             & 126.7             & 74.4k             & 7.9k              \\
            ESC@16            & 86.5 & 143.4                                  & 102.3k                                  & 16.8k                                & 63.3 & 191.5                & 136.5k                & 22.5k                & 53.0 & 223.0                & 155.5k                & 25.7k                & 68.7 & 157.7                & 66.8k                & 12.7k                \\
            SAC@16            & 76.7                                & 93.7                                & 92.4k                               & 10.0k                             & 46.7             & 122.6             & 121.1k            & 12.5k             & 39.4             & 120.1             & 116.5k            & 12.4k             & 69.7 & 60.9              & 56.6k             & 6.8k              \\
            DeepConf-high@16  & 83.3                                  & 479.6                                  & 154.5k                                  & 154.5k                                & 62.5 & 1005.6                & 179.2k                & 179.2k                & 48.5               & 767.9                & 197.6k                & 197.6k                & 70.2 & 427.3                & 119.0k                & 119.0k                \\
            DeepConf-low@16   & 76.7                                & 246.2                               & 171.2k                              & 171.2k                                & 50.0 & 255.0             & 174.3k            & 174.3k                & 42.4 & 518.7             & 179.9k            & 179.9k                & 69.2               & 163.2                & 117.0k                & 117.0k                \\
            Parallel-Probe@16 & 81.7 & 76.9                       & 85.4k                               & 8.5k                              & 60.0 & 95.6     & 98.8k             & 10.6k             & 47.7 & 93.9     & 106.0k            & 9.7k              & 67.3 & 57.4  & 61.4k             & 6.5k              \\
            \approach{}@16    & 86.7 & 79.3                    & 96.4k                               & 8.0k                              & 63.3 & 108.8 & 124.0k            & 10.8k             & 51.5 & 106.9 & 126.1k            & 10.3k             & 70.4 & 56.8     & 62.8k             & 6.5k              \\
            \bottomrule
        \end{tabular}
    \caption{Main results across four benchmarks. Acc., Lat., Tok., and Seq. denote accuracy (\%), wall-clock latency (s), total generated tokens, and sequential generated tokens, respectively.}
    \label{tab:main}
\end{table*}
 
\noindent\textbf{Benchmarks.}
We evaluate \approach on two categories of challenging reasoning benchmarks.
For competition mathematics, which demands long-horizon symbolic derivation, we employ AIME 2026~\citep{aime26}, HMMT November 2025, and HMMT February 2026~\citep{dekoninck2026matharena}.
For general scientific reasoning, which demands knowledge-intensive multi-step inference over multiple-choice options, we use GPQA Diamond~\citep{rein2024gpqa}.


\paragraph{Models.}
We use Qwen3.5-35B-A3B~\citep{qwen3.5} and GPT-OSS-20B~\citep{openai2025gptoss120bgptoss20bmodel}, two long-chain-of-thought reasoning models with different parameter scales and post-training configurations.

\paragraph{Baselines.}
We compare \approach with representative methods for parallel reasoning and adaptive self-consistency.
Baselines include \textbf{Zero-shot} with a single reasoning trajectory, \textbf{Self-consistency (SC)} with fixed-budget parallel sampling and majority voting~\citep{wang2022selfconsistency}, \textbf{ESC} with online answer aggregation and early stopping~\citep{li2024esc}, \textbf{SAC} with answer-convergence-based termination~\citep{liu2025answerconvergence}, \textbf{DeepConf} with confidence-guided trajectory filtering~\citep{fu2025deepconf}, and \textbf{Parallel-Probe} with intermediate-answer probing for consensus control and branch pruning~\citep{zheng2026parallelprobe}.

\paragraph{Metrics.}
Following prior work~\citep{zheng2026parallelprobe,fu2025deepconf}, we report accuracy (Acc.), wall-clock latency (Lat.), total generated tokens (Tok.), and sequential generated tokens (Seq.). Wall-clock latency captures end-to-end inference time, while sequential generated tokens measures critical-path computation.

\paragraph{Implementation Details.}
All experiments are conducted using vLLM~\citep{kwon2023efficient} on a single NVIDIA A100 80GB GPU. 
Unless otherwise specified, we use a maximum generation budget of 16,384 tokens per trajectory, nucleus sampling with temperature 0.6 and $p=0.95$, and $K=16$ parallel branches for all methods.
For \approach, probes are issued every $\tau=500$ generated tokens and retain the top-$L=20$ answer candidates. 
The default configuration sets $W=7$, $X=9$, $N_{\mathrm{warm}}=15$, $q_{\mathrm{prune}}=0.50$, $\theta_{\mathrm{retire}}=0.90$, and $\gamma_{\mathrm{ES}}=0.50$. 
All reported results are averaged over four independent runs.

\subsection{Main Results}

Table~\ref{tab:main} summarizes the performance of \approach and baselines across four reasoning benchmarks.
\approach consistently achieves an effective accuracy--efficiency trade-off, reducing inference cost while preserving the benefits of parallel reasoning.
Compared with self-consistency (SC), \approach maintains competitive accuracy with substantially lower latency and token consumption.
For Qwen3.5-35B-A3B, \approach achieves 71.1\% average accuracy, within 1.1 percentage points of SC (72.2\%), while reducing latency and total generated tokens by 21.8\% and 30.3\%, respectively.

Among fully parallel controllers, \approach significantly improves the accuracy--efficiency balance.
SAC reduces computation by terminating when sampled answers agree, but its stopping criterion does not explicitly account for branch-level progress.
In contrast, \approach reallocates computation according to branch convergence.
Compared with Parallel-Probe, which relies on synchronized cross-branch consensus, \approach improves accuracy by 3.9 points while reducing latency by 10.6\% on Qwen3.5-35B-A3B; on GPT-OSS-20B, it achieves a 3.8-point accuracy improvement with comparable sequential cost.

We further compare \approach with controllers involving sequential or partially sequential execution.
DeepConf and ESC reduce unnecessary sampling through confidence-based decisions, but their control flow introduces sequential dependencies that increase critical-path cost.
By performing branch-level decisions asynchronously during generation, \approach achieves comparable or higher accuracy with substantially lower latency.

\begin{figure}[t]
    \centering
    \includegraphics[width=0.9\columnwidth]{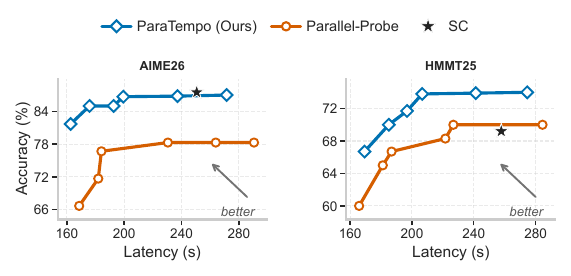}
    \caption{Latency--accuracy scaling curves of \approach and Parallel-Probe with Qwen3.5-35B-A3B. The star marks self-consistency (SC@16 in Table~\ref{tab:main}). \approach achieves a superior Pareto frontier across different operating points.}
    \label{fig:pspp-pareto}
\end{figure}

Figure~\ref{fig:pspp-pareto} presents the latency--accuracy trade-off across controller configurations.
On AIME26 and HMMT25, the \approach frontier consistently matches or exceeds that of Parallel-Probe across the evaluated latency range: under similar latency budgets, \approach achieves higher accuracy, and additional computation continues to improve performance before saturation.
Moreover, the highest-budget \approach configurations achieve accuracy comparable to or higher than SC@16 while reducing latency by approximately 20\%.
These results demonstrate that \approach provides robust improvements across different operating points.

\begin{figure}[t]
    \centering
    \includegraphics[width=\columnwidth]{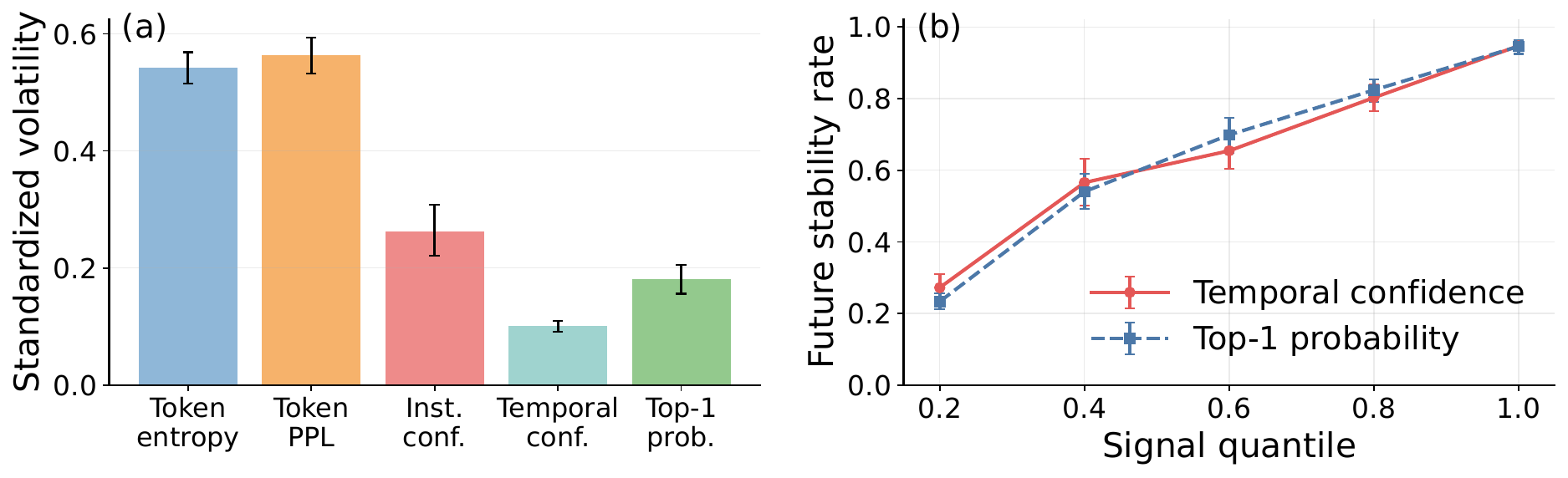}
    \caption{Analysis of temporal confidence under the preliminary study protocol. 
    (a) Temporal confidence exhibits lower volatility than token-level signals and instantaneous answer confidence. (b) The future stability rate ($h=5$) increases monotonically with signal quantiles of temporal confidence and top-1 probability. 
    }
    \label{fig:prelim}
\end{figure}

\subsection{Temporal Confidence Analysis}
\label{sec:tempconf-analysis}

We further examine whether temporal confidence serves as an effective signal for online branch control. 
Following the evaluation protocol in the preliminary study (Section~\ref{sec:prelim}), we evaluate two key properties of a control signal: temporal stability and future convergence prediction.
Temporal confidence and top-1 probability are computed with an aggregation window of $W{=}5$ throughout this analysis.

\paragraph{Temporal Stability.}
Figure~\ref{fig:prelim}(a) compares the volatility of temporal confidence with the signals analyzed in the preliminary study. 
Temporal confidence produces substantially smoother trajectories than token-level uncertainty measures and instantaneous answer confidence. 
This indicates that temporal aggregation effectively suppresses transient fluctuations and provides a more reliable basis for branch-level decisions.

\paragraph{Future Convergence Prediction.}
Beyond stability, an effective control signal should provide predictive information about future branch behavior. 
Figure~\ref{fig:prelim}(b) reports empirical future-stability rates over signal quantiles of temporal confidence and top-1 probability.
Both measures show consistent positive correlations with future stability: branches with higher values are more likely to preserve their dominant answers over subsequent probes. 
These results demonstrate that temporal confidence captures meaningful convergence trends and supports the adaptive control decisions in \approach.


\subsection{Ablation Study}


\approach performs adaptive computation allocation through three temporal-confidence-driven operations: pruning, retirement, and forking. We evaluate their contributions through individual ablations: \textbf{w/o Prune} disables pruning, \textbf{w/o Retire} removes early retirement, and \textbf{w/o Fork} disables budget reallocation via forking.

Table~\ref{tab:ablation} reports results on HMMT25, with extended results on AIME26 and HMMT26 provided in the Technical Supplement. Removing pruning increases latency with only a marginal accuracy change, demonstrating that pruning effectively eliminates unpromising branches. Removing retirement increases total tokens and latency, while reducing accuracy by $6.6$ points, showing that early retirement avoids redundant decoding and preserves reliable voting evidence from converged branches. Disabling forking achieves the lowest computation cost but reduces accuracy by $3.3$ points, indicating that reusing released computation for exploration is important for maintaining solution diversity. Overall, the complete \approach configuration achieves the best accuracy while retaining substantial efficiency gains, confirming that pruning, retirement, and forking provide complementary benefits.

\begin{table}[t]
    \centering
    \setlength{\tabcolsep}{9pt}
    \begin{tabular}{lcccc}
        \toprule
        \textbf{Variant} & \textbf{Acc $\uparrow$} & \textbf{Lat $\downarrow$} & \textbf{Tok $\downarrow$} & \textbf{Seq $\downarrow$} \\
        \midrule
        \approach        & 73.3           & 205.7                     & 166.6k                    & 10.6k                     \\
        \quad w/o Prune  & 71.7                    & 233.0                     & 179.1k                    & 11.4k                     \\
        \quad w/o Retire & 66.7                    & 224.6                     & 175.4k                    & 10.4k            \\
        \quad w/o Fork   & 70.0                    & 195.7            & 145.5k           & 10.4k            \\
        \bottomrule
    \end{tabular}
    \caption{Ablation results of \approach on HMMT25 with Qwen3.5-35B-A3B. Acc., Lat., Tok., and Seq. denote accuracy (\%), latency (s), total tokens, and sequential tokens (K).  }
    \label{tab:ablation}
\end{table}

\section{Related Work}
\paragraph{Efficient Test-Time Scaling.}
Test-time scaling improves reasoning by allocating additional inference computation to longer trajectories or diverse sampled solutions~\citep{wang2022selfconsistency,snell2024scaling,zeng2026dockerless,gao2026swe,lin2026knowfixqadrivenrepository}. 
Self-consistency scales inference width by sampling multiple reasoning paths and aggregating final answers, but fixed rollout budgets ignore variation in problem difficulty and branch utility. 
Recent methods improve sample efficiency through adaptive sampling, weighting, and allocation. 
ESC stops sampling when answer agreement is sufficient~\citep{li2024esc}, DSC adjusts rollout budgets based on problem difficulty~\citep{wang2025dsc}, and RASC, CISC, RPC, and DORA further exploit rationale quality, confidence signals, or rollout allocation strategies~\citep{wan2025rasc,taubenfeld2025cisc,zhou2025rpc,wang2025dora,ma2026llmagentscoderepositories,hu2026line,chen2025swe}. 
However, these approaches primarily operate at the sampling or trajectory level and provide limited control over branches during generation. 
In contrast, \approach treats parallel reasoning as an online branch-level control problem and adapts computation allocation while trajectories are still evolving.

\paragraph{Parallel Reasoning.}
Parallel reasoning improves inference robustness by exploring multiple reasoning trajectories concurrently, ranging from skeleton-based expansion~\citep{ning2023skeleton} to adaptive trajectory scaling and pruning~\citep{shi2025longcodezip,shi2024code,li2025swe,zhang2026swe}. 
Slim-SC removes redundant chains via online inter-trace similarity~\citep{hong2025slimsc}, while DeepPrune clusters partial traces using predicted answer equivalence~\citep{tu2026deepprune}. 
Another line of work introduces intermediate signals for early control: Answer Convergence tracks answer stability~\citep{liu2025answerconvergence}, DeepConf uses token-level confidence for trajectory filtering~\citep{fu2025deepconf}, and TRACE aggregates temporal consistency for early exit~\citep{li2026trace}. 
Most closely related, Parallel-Probe performs synchronized intermediate probing for cross-branch consensus control~\citep{zheng2026parallelprobe}, while ATTS reduces synchronization overhead through speculative scaling~\citep{xiong2025atts}. 
In contrast, \approach uses temporally aggregated answer distributions as a branch-local signal, enabling asynchronous lifecycle control through pruning, retirement, forking, and confidence-weighted voting.




\section{Conclusion}
In this work, we introduced \approach, a training-free asynchronous framework for efficient parallel test-time reasoning. 
Central to \approach is \textit{temporal confidence}, a branch-local convergence signal that captures answer evolution through temporally aggregated intermediate answer distributions.
By leveraging this signal, \approach enables adaptive branch-level computation allocation through pruning, retirement, exploration, and confidence-weighted termination.
Experiments on challenging reasoning benchmarks demonstrate that \approach significantly reduces inference cost while preserving the accuracy and robustness of parallel reasoning.

\bibliography{references}

\end{document}